\documentclass[sigconf]{acmart}

\AtBeginDocument{%
  \providecommand\BibTeX{{%
    \normalfont B\kern-0.5em{\scshape i\kern-0.25em b}\kern-0.8em\TeX}}}

\setcopyright{acmcopyright}
\copyrightyear{2018}
\acmYear{2018}
\acmDOI{10.1145/1122445.1122456}

\acmConference[Woodstock '18]{Woodstock '18: ACM Symposium on Neural
  Gaze Detection}{June 03--05, 2018}{Woodstock, NY}
\acmBooktitle{Woodstock '18: ACM Symposium on Neural Gaze Detection,
  June 03--05, 2018, Woodstock, NY}
\acmPrice{15.00}
\acmISBN{978-1-4503-XXXX-X/18/06}

\acmSubmissionID{2502}

\usepackage{amsthm}
\usepackage{amsmath}
\usepackage{caption}
\usepackage{subfigure}
\usepackage{graphicx}
\usepackage{hyperref}
\makeatletter
\def\UrlAlphabet{%
      \do\a\do\b\do\c\do\d\do\e\do\f\do\g\do\h\do\i\do\j%
      \do\k\do\l\do\m\do\n\do\o\do\p\do\q\do\r\do\s\do\t%
      \do\u\do\v\do\w\do\x\do\y\do\z\do\A\do\B\do\C\do\D%
      \do\E\do\F\do\G\do\H\do\I\do\J\do\K\do\L\do\M\do\N%
      \do\O\do\P\do\Q\do\R\do\S\do\T\do\U\do\V\do\W\do\X%
      \do\Y\do\Z}
\def\UrlDigits{\do\1\do\2\do\3\do\4\do\5\do\6\do\7\do\8\do\9\do\0}
\g@addto@macro{\UrlBreaks}{\UrlOrds}
\g@addto@macro{\UrlBreaks}{\UrlAlphabet}
\g@addto@macro{\UrlBreaks}{\UrlDigits}
\makeatother

\begin{document}

\title{Arbitrary Distribution Modeling with Censorship in Real-Time Bidding Advertising}

\author{Xu Li}
\affiliation{%
  \institution{FreeWheel}
  \city{Beijing}
  \country{China}
}
\email{lixu@apac.freewheel.com}

\author{Michelle Ma Zhang}
\affiliation{%
    \institution{Northwestern University}
    \city{Evanston}
    \state{Illinois}
    \country{USA}
}
\email{michellezhang2023@u.northwestern.edu}

\author{Youjun Tong}
\email{yjtong@apac.freewheel.com}
\author{Zhenya Wang}
\email{zywang@apac.freewheel.com}
\affiliation{%
  \institution{FreeWheel}
  \city{Beijing}
  \country{China}
}

\renewcommand{\shortauthors}{Xu and Youjun, et al.}

\begin{abstract}
  The purpose of Inventory Pricing is to bid the right prices to online ad opportunities, which is crucial for a Demand-Side Platform (DSP) to win advertising auctions in Real-Time Bidding (RTB). In the planning stage, advertisers need the forecast of probabilistic models to make bidding decisions. However, most of the previous works made strong assumptions on the distribution form of the winning price, which reduced their accuracy and weakened their ability to make generalizations. Though some works recently tried to fit the distribution directly, their complex structure lacked efficiency on online inference. In this paper, we devise a novel loss function, Neighborhood Likelihood Loss (NLL), collaborating with a proposed framework, Arbitrary Distribution Modeling (ADM), to predict the winning price distribution under censorship with no pre-assumption required. We conducted experiments on two real-world experimental datasets and one large-scale, non-simulated production dataset in our system. Experiments showed that ADM outperformed the baselines both on algorithm and business metrics. By replaying historical data of the production environment, this method was shown to lead to good yield in our system. Without any pre-assumed specific distribution form, ADM showed significant advantages in effectiveness and efficiency, demonstrating its great capability in modeling sophisticated price landscapes.
\end{abstract}



\maketitle

\section{Introduction}
\begin{figure}[ht]
    \centering
    \includegraphics[width=0.85\linewidth]{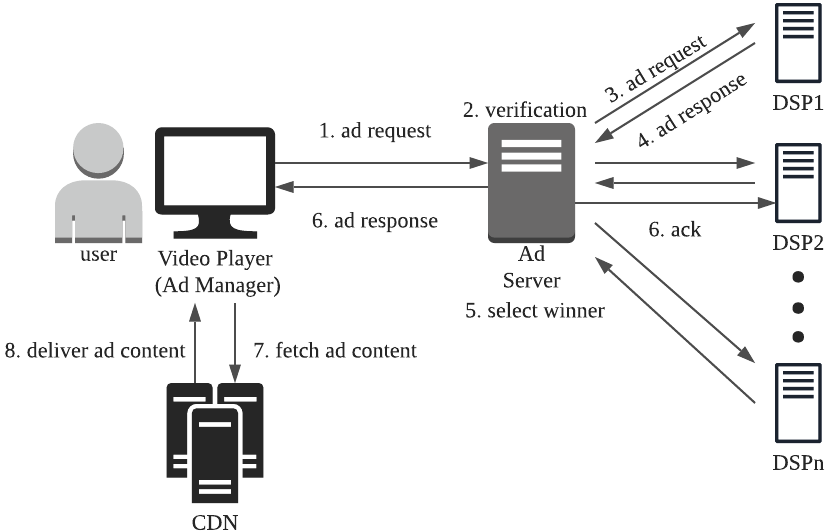}
    \caption{The Real-Time Bidding workflow (The enumeration for the arrows indicates the order of execution of the ad delivery process).}
    \Description{The Real-Time Bidding workflow.}
    \label{fig:rtb}
\end{figure}

Under the background of Digital Media, E-commerce, and Social Network Services, computational advertising plays an important role in the profits of Internet companies. Real-Time Bidding (RTB) is one of the most common scenarios in online advertising. The workflow of RTB is shown in Figure~\ref{fig:rtb}. When a user is watching a video on a website and the video comes across an opportunity to present an ad with no preset ad campaign, the ad manager of the video player will collect information about the inventory and send a request to the ad server. The ad server verifies the ad request and raises an auction, forwarding the ad request with necessary information to multiple buyers, which are Demand-Side Platform (DSP). The DSP, on behalf of the client advertisers, may separately return multiple bid responses with proper bidding prices for different advertisers based on their preferences. The ad server verifies the validity of the candidates based on a series of rules and finally chooses the one with the highest bidding price as the winner. The ad server then sends an acknowledgment back to the winner through the DSP and sends a response to the ad manager. The ad content is finally fetched and delivered from Content Distribution Network (CDN) by the address in the response to the video player and displayed on the screen before the users' eyes.

Predicting the winning price, as known as Inventory Pricing, is the key to winning the RTB auction for a DSP \cite{overview1, overview2}. Previous works have tried to use Survival Analysis (SA) \cite{sa} or machine learning algorithms \cite{mixture1, kdd21neural_auction} to forecast the winning price. These methods provide a single price value as a recommendation for the inventory price, and so they are called the Point Estimator. However, in the business, a DSP may need more information to help make pricing decisions, for example, ``how much I should bid if I want to win a certain amount of impressions?''. This can be calculated by multiplying the winning rate of certain prices and the predicted gross inventory of the specific dimensions. As such, since the point estimators failed to provide price landscape information, probabilistic models were adapted to this task to obtain informative prediction results. Unfortunately, most solutions make an assumption about the winning price distribution form beforehand, which is a strict presupposition that can seriously impact the prediction accuracy. Previous works \cite{dlf, kdd21dist} and the observations in our system both demonstrate that the distribution of the winning price is not subjected to one specific classic distribution form. Moreover, the winning price distribution of each ad impression varies significantly and it's impossible to choose one common distribution form to fit them all. Though some works \cite{dlf, kdd21meituan} have tried to fit the price landscape without any pre-assumption, their complex structures increase inference time, which is still a concern for the advertising system that requires instantaneous responses.

Meanwhile, the applications of the deep learning method on this task were mainly focused on proposing novel prediction functions to fit the landscape, overlooking the role of feature extraction in profiling the ad requests. Various 2-order feature extraction components in previous works \cite{innerprod, fm, afm, dcn-mix, cin, autoint, nfm} have proved their effectiveness. In the context of RTB, the 2-order feature could be considered a sort of Collaborative Filtering, which indicates the value of the ad impression by measuring the relativity of the website features and the user features. For example, the bidding price of a Nike gear ad to an impression of a male teenager who's watching an NBA video should be higher than that of an elderly person in the same sports site section. We deem that the 2-order feature could not be determined simply by a linear combination of 1-order features. The deep structure of the neural network has advantages in profiling the bid requests and fitting sophisticated landscapes.

\begin{figure}[ht]
    \centering
    \includegraphics[width=\linewidth]{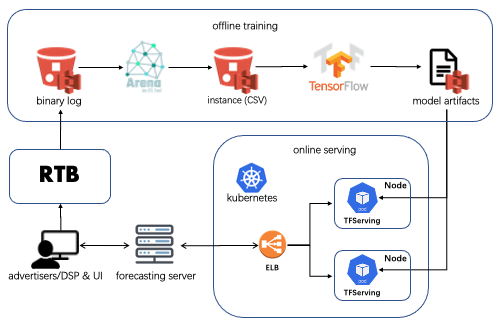}
    \caption{The system architecture of Inventory Planning.}
    \Description{The system architecture of Inventory Planning.}
    \label{fig:sys}
\end{figure}

Figure~\ref{fig:sys} shows a brief architecture of our system. The training samples are extracted from Binary Log by ArenaETL, which is an internal presto-based ETL (Extract, Transform, Load) tool. Data cleaning was also executed in this stage. For the instance of missing values, we filled in -1 for id, 0 for metrics and '' for string type fields. As for the feature pre-processing, we left it to be processed by the feature column layers in the model. Then, we train the model with TensorFlow and save the model artifacts on S3. TensorFlow Serving can scan, upload and deploy the model under a specific directory automatically. At the Inventory Planning stage, the requests are sent to Kubernetes when DSP or advertisers ask for the winning price landscape of a set of dimensions, where Elastic Load Balancing (ELB) balance the stress and distribute them to each node where TensorFlow Serving is running. These infrastructures guarantee a Service-Level Agreement (SLA) of 10ms. With the predicted distribution of winning price as reference, the DSP or advertisers can then make decisions on pricing and book campaigns to participate in RTB, which works programmatically as the process in Figure~\ref{fig:rtb}, which is omitted here. Finally, the logs are collected into Binary Log. As a newly released feature of our product, we replayed historical data in our system to measure its performance. And since there is no essential difference between historical and future samples, performance on historical data is enough for us to prove its effectiveness.

In this paper, we proposed a novel optimization objective with a concise framework. And a sufficient set of experiments was conducted to verify the effectiveness and efficiency. The main contributions of this paper are:
\begin{enumerate}
    \item we devised a more accurate loss function, Neighborhood Likelihood Loss (NLL), to model arbitrary distribution without any prior distribution form assumption. Its effectiveness has been verified on different structures.
    \item we devised a concise and extensible neural network framework, Arbitrary Distribution Modeling (ADM), which is simple but efficient and suitable for advertising online serving.
    \item Meanwhile, besides common algorithm metrics, for the first time we leveraged some business metrics that advertisers care about in the RTB scenario in comparing different predicting baselines, which demonstrates the yield advantage of ADM against the others.
\end{enumerate}
The implementation of the codes is available on \href{https://github.com/shinleylee/Arbitrary_Distribution_Modeling.git}{github}.

The rest of this paper is organized as follows. In the next section, we will give a brief overview of the traditional and recently proposed methods in Inventory Pricing. In section \pageref{page:methodology}, we will illustrate the loss function as well as the proposed framework in detail. The experiment settings and results is presented in section \pageref{page:experiments} and section \pageref{page:evaluation} respectively. Finally, we draw the conclusion in the last section.

\section{Related Works}
\label{page:related_works}

The traditional method to solve this task was Survival Analysis, a branch of statistics for analyzing the expected duration of time until one event occurs, most commonly used in the medical field \cite{sa}. Some machine learning models, like LR \cite{mixture1}, GBDT \cite{gbdt}, were leveraged to make better predictions. However, as mentioned above, the point estimators only provide a value for the winning price and fail to give more information about the price distribution to help DSP make better bidding decisions.

Probabilistic estimators were later introduced into Inventory Pricing in RTB \cite{mixture2, gamma1, ass1, ass2, ass3, beeswax}. They set an assumption for the distribution form of the winning price and use models to fit the key parameters in Gaussian \cite{mixture1, dlf}, Log-Normal \cite{gbdt}, Gamma \cite{gamma1, gamma2}, or other parametric distributions \cite{ass2, ass3, beeswax}. This allows the DSP to acquire the landscape of the winning price and be able to include more strategy when deciding the bidding price rather than directly bidding the predicted result from a point estimator. However, the inherent defect of these pre-assumed probabilistic models is that the distribution assumption could severely lessen the model's effectiveness. The appropriate price distribution form may vary for different datasets, and even for different ad dimensions. It is almost impossible to use a fixed distribution assumption to accurately describe the price landscape. Meanwhile, our system and some works \cite{dlf, kdd21dist} also observed that the distribution of the winning price is too complex to fit in any specific classic distribution.

Recently, some non-assumption probabilistic models were proposed. Deep Landscape Forecasting (DLF) is one that leverages Recurrent Neural Network (RNN) to fit the conditional probabilities to model the distribution of the winning price with no prior assumption required. It reached a state-of-the-art performance on modeling sophisticated distributions \cite{dlf}. Unfortunately, to guarantee the precision of the prediction, the price range has to be split into trivial intervals, which leads to hundreds of time steps for the Long Short-Term Memory (LSTM), weakening its efficiency. Also, some other works introduced more complex structures \cite{dlf2, kdd21meituan} to fit the arbitrary landscape, which leads to the increase of time cost on online inference. Both the temporal dependency or the complex structure models are not friendly to the ad server system, which requires millisecond-level responses with concurrency. We need a solution to fit the price landscape accurately with low time overhead on online serving.

\section{Methodology}
\label{page:methodology}

\subsection{Preliminaries}
First, we will introduce the concept of the ``Second Price'' principle. When the ad server collects all of the bidding prices from the DSP, it first verifies the responses and figures out who is the winner with the highest bidding price. Then, the winner is charged the second-highest bidding price for this ad opportunity. Therefore, the target we are going to predict is the lowest possible price to bid to win the auction.

Given this mechanism, the concept of ``Censorship'' is introduced into RTB. When a DSP wins the auction, it will know the winning price of this auction from the acknowledgment response of the ad server. However, when a DSP loses the auction, it receives no response from the ad server and thus has no idea how high the winning price was. All the DSP knows is that the winning price was higher than their bidding price. This scenario of censored data is called ``Censorship'' \cite{censorship1, censorship2}.

There have been studies demonstrating the importance of involving the censored data into model training \cite{mixture1, mixture2}. Only using the uncensored winning samples to fit a regression model will always lead to a lower prediction. In an auction, the winning price for the winning DSP is the second-highest price of this auction. On the other hand, for a losing DSP, the winning price is the highest price of this auction - they had to have bid higher than the winning DSP to win the auction. The difference of the distribution of the winning price between censored and uncensored data has been found theoretically and practically in previous works \cite{mixture1, mixture2}, and using a combination of both sets of data has been verified to be most effective. General practice to make use of these two sets of data is to fit a probability density function (P.D.F.) of the winning price from the winning samples, and to fit a cumulative distribution function (C.D.F.) of the bidding price from the losing samples, where the C.D.F. can be understood as the winning rate. This is achieved by optimizing the log-likelihood loss function \cite{dlf}.

\subsection{Problem Definition}

In the actual RTB scenario, a DSP is finally chosen as the winner not only depending on its bidding price, but also from business measurements such as the frequency cap, budget pacing, competitor constraint, validity, legality, etc. In this paper, we only study the prediction of the winning price under the simplified scenario where the price is the only condition to select the winner of an RTB auction.

Briefly, the problem we want to solve is to predict a distribution of the winning price from various features of the ad request. There are four types of features in deciding the distribution of the winning price:
\begin{itemize}
    \item The features of the publisher $\boldsymbol{x}_p$ is the information about the website, which includes the site URL, domain, video group, ad slot size, ad min/max duration, etc; 
    \item The features of the user $\boldsymbol{x}_u$ is the information about the viewer, which includes the gender, age, location, device, platform, etc; 
    \item The features of the ad $\boldsymbol{x}_a$ is the information about the ad DSP wants to deliver, which includes the advertiser id, industry, genre, ad duration, etc. $\boldsymbol{x}_a$ is not shareable and different advertisers have their unique ad features. 
    \item The features of the context $\boldsymbol{x}_c$ is the neutral information like the request timestamp, week of day, daypart, etc.
\end{itemize}

For the \emph{ad requests} $\boldsymbol{x} = \{\boldsymbol{x}_p, \boldsymbol{x}_u, \boldsymbol{x}_a, \boldsymbol{x}_c\}$, we denote \emph{the winning price} as $z$ and denote \emph{the bidding price} as $b$. We define \textbf{the distribution of the winning price} $pr_z(*)$ as the P.D.F. of any price value $p$ to be the winning price
\begin{equation}
    \label{equ:pdf}
    pr_z(p|\boldsymbol{x}) \triangleq P(z=p|\boldsymbol{x})
\end{equation}
where $P(*)$ is the landscape of the probability on the whole ad feature space $\boldsymbol{\mathcal{X}}$, and $\boldsymbol{x} \in \boldsymbol{\mathcal{X}}$. This is the prediction target of the probabilistic methods of the Inventory Pricing task. On the other hand, we define the probability of one bidding price $b$ to win the auction as the \textbf{winning rate}, $wr(b)$, i.e. it indicates the proportion of multiple independent identical repeat ad requests $\boldsymbol{x}$ that $b$ could win. This could be calculated by the C.D.F of $pr_z(p|x)$ as Equation~\ref{equ:wr}, 
\begin{align}
    \label{equ:wr}
    wr(b) &= \int_{p^l}^{b} pr_z(p|\boldsymbol{x}) \mathrm{d}p  \\
    &\approx \sum_{i=i_{p^l}}^{i_b} pr_i(i|\boldsymbol{x})
\end{align}
where $p^l$ is the left boundary of the price range, i.e. the lowest price for all $\boldsymbol{x}$ in $\boldsymbol{\mathcal{X}}$. The C.D.F. guarantees higher price has higher chance to win (winning rate), but it doesn't necessarily mean that it's more likely to be the winning price (probability in P.D.F.).

The boundaries of the price range should be set by the statistics of the dataset accordingly. Because predicting a continuous distribution function as output is impractical for a model, we discrete the price range into buckets as previous works\cite{dlf, dlf2} did. The width of each bucket is determined by the requirement of precision in business. The winning rate becomes a sum of the probabilities of the buckets in range as the approximation part in Equation~\ref{equ:wr}, where $i_b$ denotes the bucket index of the bidding price and $pr_{\boldsymbol{i}}$ denotes the discrete price distribution density function. In this way, we transfer this problem into a classification task.

\subsection{Arbitrary Distribution Modeling}

We propose an \textbf{Arbitrary Distribution Modeling} (ADM) framework, which makes no assumption of the distribution form to forecast the price landscape. Its loss function helps learn the accurate landscape and its simple structure guarantees its efficiency. To leverage this framework, we first separate the price range into $N$ equal-length price intervals and then try to predict the probability of the winning price of each interval.

\subsubsection{Loss Function}

We define the \textbf{neighborhood} of $z$ as a range where the difference between the prices and $z$ is less than a relatively small distance $\delta$. Similarly, we define the left neighborhood and the right neighborhood as Equation~\ref{equ:neighborhood}
\begin{equation}
\begin{aligned}
\label{equ:neighborhood}
U(z,\delta) &= (z - \delta, z + \delta) = \{ p | |z - p| < \delta \}\\
U^{-}(z,\delta) &= (z - \delta, z] = \{ p | 0 \leq z - p < \delta \}\\
U^{+}(z,\delta) &= [z, z + \delta) = \{ p | 0 \leq p - z < \delta \}
\end{aligned}
\end{equation}

It is reasonable to assume that, to win the auction, the bidding prices from DSP should be close to the winning prices in general. This is also verified empirically in our system as Figure~\ref{fig:assumption} shows:
\begin{figure}
    \centering
    \subfigure[The distribution of the difference between $b$ and $z$.]{
        \includegraphics[scale=0.39]{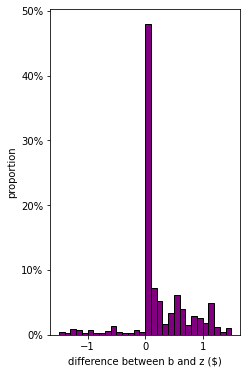}}
        \label{fig:diff_dist}
    \vspace{0.5cm}
    \subfigure[The distribution of the bidding price $b$ and the winning price $z$.]{
        \begin{minipage}[b]{0.45\linewidth}
        \includegraphics[scale=0.35]{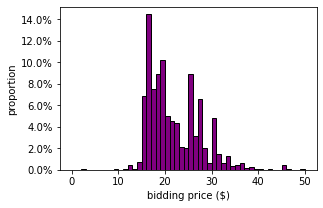}
        \includegraphics[scale=0.35]{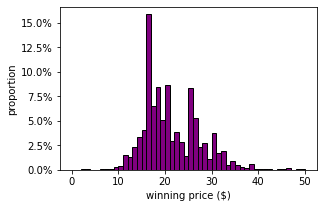}
        \end{minipage}}
        \label{fig:bz_dist}
    \caption{The distributions of $b$, $z$ and their differences.}
    \label{fig:assumption}
\end{figure}
over $58\%$ of all the winning and losing transactions, the difference between $b$ and $z$ is within $\pm0.1$ dollar and over $93\%$ no more than 1 dollar, while most price values fall in between the tens of dollars. So, we assumed
\newtheorem{assumption}{Assumption}
\begin{assumption}
    The bidding price $b$ is within the neighborhood of the winning price $z$ for both winning and losing bids, i.e. $b \in U(z,\delta)$.
    \label{assumption}
\end{assumption}

Based on this assumption, we proposed the \textbf{Neighborhood Likelihood Loss} (NLL), which provides a more precise direction than the previous ones in guiding the model learning from the observations. From the perspective of a DSP, the observations are $(\boldsymbol{x}^k, b^k, z^k)$ from the winning sets $\mathbb{D}_{win}$ and $(\boldsymbol{x}^k, b^k, ?)$ from the losing sets $\mathbb{D}_{win}$, where $k$ is the identifier of different bids. The core idea of this loss is to promote the right probabilities accurately according to the observations. It consists of three parts:

First, we devised a P.D.F. loss to maximize the probability of the winning prices for the winning bids in $\mathbb{D}_{win}$. This is implemented by minimizing the negative log-likelihood
\begin{equation}
\begin{aligned}
Loss_1 &= -\log\prod_{(\boldsymbol{x}^k,z^k)\in\mathbb{D}_{win}} pr_z(z^k|\boldsymbol{x}^k)\\
&=-\log\prod_{(\boldsymbol{x}^k,z^k)\in\mathbb{D}_{win}}pr_{\boldsymbol{x}^k}({i_{z^k}})\\
&=-\sum_{(\boldsymbol{x}^k,z^k)\in\mathbb{D}_{win}}\log pr_{\boldsymbol{x}^k}({i_{z^k}})\\
\end{aligned}
\end{equation}

\begin{figure}[ht]
\centering
    \subfigure[The probabilities of prices in the neighborhood of $z^k$ need to be promoted for samples in $\mathbb{D}_{win}$.]{
    \centering
        \includegraphics[scale=0.18]{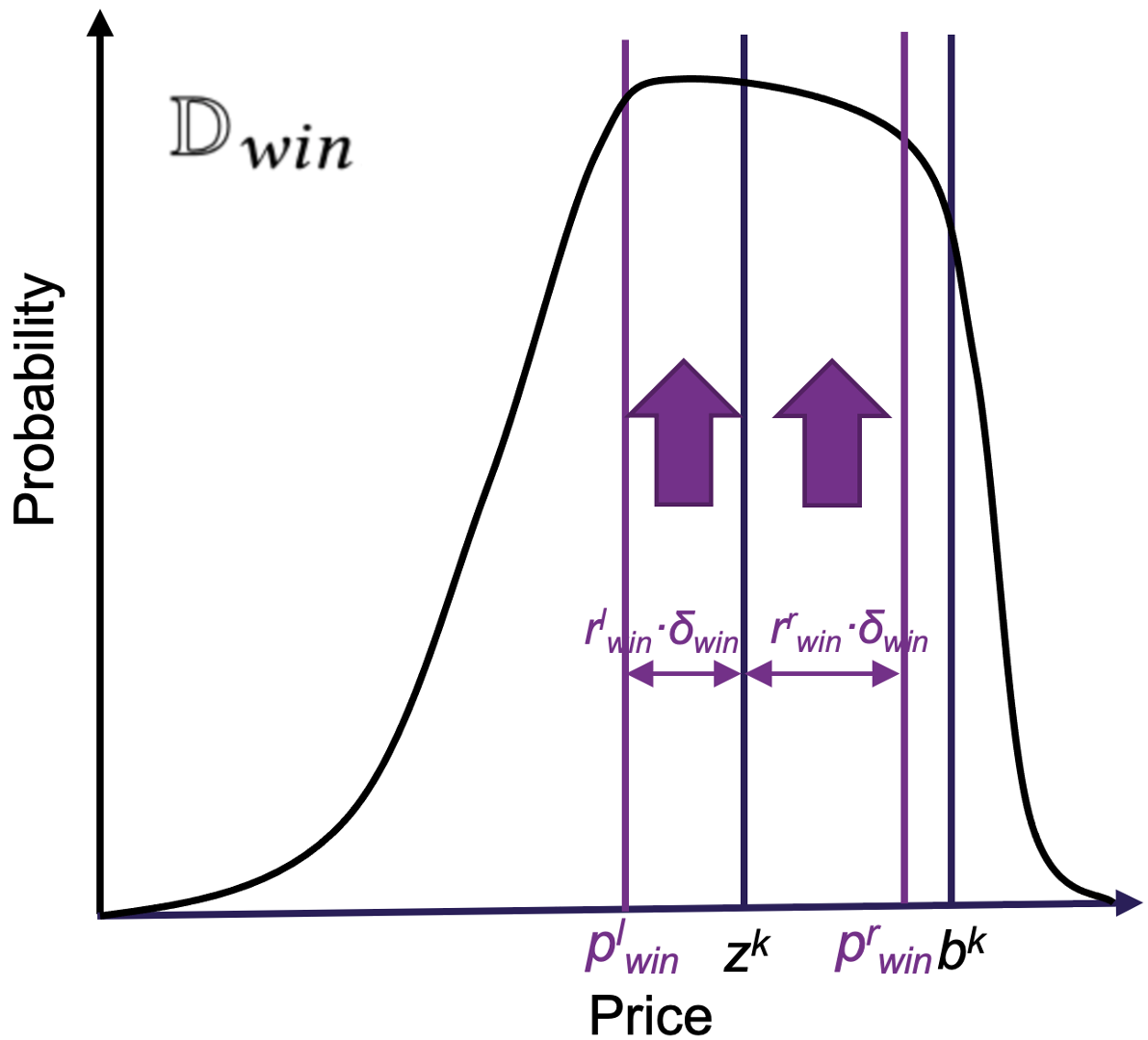}
        \label{fig:zone_win}}
    \subfigure[The probabilities of prices in the right neighborhood of $b^k$ need to be promoted for samples in $\mathbb{D}_{lose}$.]{
    \centering
        \includegraphics[scale=0.18]{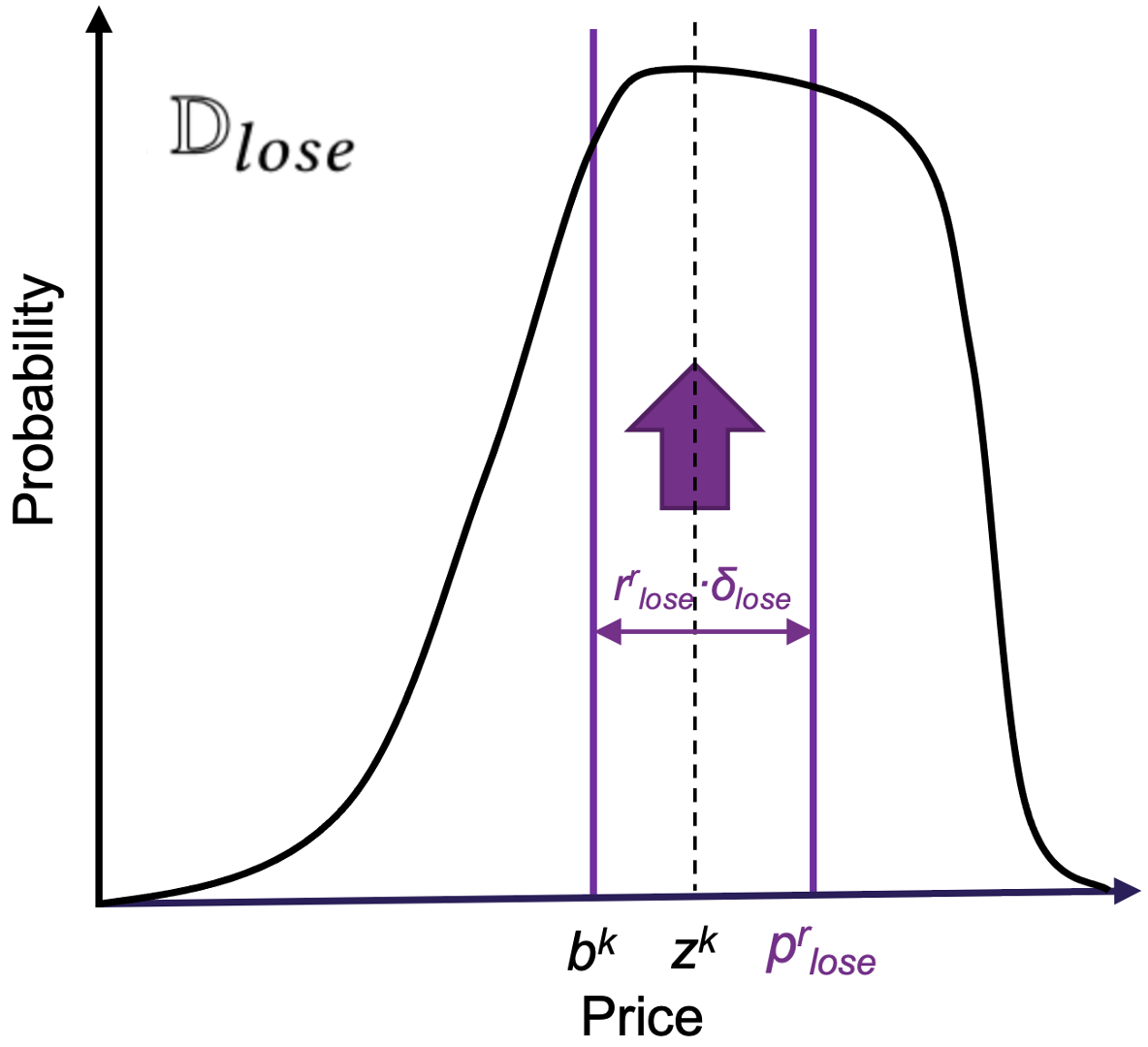}
        \label{fig:zone_lose}}
\caption{The optimization zones in $\mathbb{D}_{win}$ and $\mathbb{D}_{lose}$.}
\Description{The optimization zones in price range.}
\label{fig:zone}
\end{figure}

Second, we maximize the winning rate of the winning prices $wr(z^k)$ for the winning bids in $\mathbb{D}_{win}$. In previous works \cite{mixture2, dlf}, maximizing the Equation \ref{equ:wr} is the optimization objective for the winning rate with $p^l$ set to $0$ . This can lead to equal promotion of the probabilities for prices between $0$ and $z^k$. However, very low prices are unlikely to become the winning price and as such, their probabilities should stay low. There should be a left boundary instead of $0$ to limit the promotion range. The promotion of $wr(z^k)$ should result from the promotion of the prices within the left neighborhood $U^{-}(z^k,\delta)$. On the other hand, the premise of probabilistic Inventory Pricing is the winning price uncertainty, which means the winning price is obtained from a specific distribution rather than a fixed value each time. In other words, the winning price for an identical ad request $\boldsymbol{x}^k$ may be different next time, and it should fluctuate within the neighborhood of the observed winning price $z^k$. So, there should also be a right neighborhood with a right boundary $U^{+}(z^k,\delta)$
. Given our Assumption~\ref{assumption}, we deem that a ratio of the distance $\delta_{win} = b^k - z^k$ is an appropriate breadth for neighbourhood. Thus, we set the right boundary $ p_{win}^r = z^k + r_{win}^r \cdot \delta_{win}$, and the left boundary $p_{win}^l = z^k - r_{win}^l \cdot \delta_{win}$, where $r_{win}^r$ and $r_{win}^l$ are the hyper-parameters of ratios to control the zone breadth, as Figure~\ref{fig:zone_win}.
In this way, we have our second loss function on the winning bids $\mathbb{D}_{win}$, which is devised as
\begin{equation}
    \begin{aligned}
        Loss_2
        =& -\log\prod_{(\boldsymbol{x}^k,z^k)\in\mathbb{D}_{win}} \left[{wr(z^k|\boldsymbol{x}^k) + \int_{U^{+}(z^k,r_{win}^r\cdot\delta_{win})} pr_z(p|\boldsymbol{x}^k) \mathrm{d}p }\right] \\
        =& -\log\prod_{(\boldsymbol{x}^k,z^k)\in\mathbb{D}_{win}} \left[{\int_{p_{win}^l}^{z^k} pr_z(p|\boldsymbol{x}^k) \mathrm{d}p} + \int_{z^k}^{p_{win}^r} pr_z(p|\boldsymbol{x}^k) \mathrm{d}p\right] \\
        =& -\log\prod_{(\boldsymbol{x}^k,z^k)\in\mathbb{D}_{win}} \int_{p_{win}^l}^{p_{win}^r} pr_z(p|\boldsymbol{x}^k) \mathrm{d}p\\
        \approx& -\log\prod_{(\boldsymbol{x}^k,z^k)\in\mathbb{D}_{win}} \sum_{i=i_{p_{win}^l}}^{i_{p_{win}^r}} pr_i(i|\boldsymbol{x}^k)\\
        =& -\sum_{(\boldsymbol{x}^k,z^k)\in\mathbb{D}_{win}} \log\sum_{i=i_{p_{win}^l}}^{i_{p_{win}^r}} pr_i(i|\boldsymbol{x}^k)\\
    \end{aligned}
\end{equation}
where the left boundary of the $wr$ promotion range $p^l$ in Equation~\ref{equ:wr} is no longer $0$ or the minimum price, but the left boundary of the left neighborhood. And the losses used in previous works \cite{mixture2, dlf} are special cases when the left boundary is $0$ and the right boundary is set to the bidding price $b^k$.

Third, the only information from $\mathbb{D}_{lose}$ is that the winning price is greater than the observed bidding price. As such, some but not all of the probabilities of the prices on the right of $b^k$ should be promoted. We should note that, based on the Assumption~\ref{assumption}, very high prices are not likely to be the winning price. The prices in $U^{+}(b^k, \delta)$ should be the target whose probabilities need be promoted, as shown in Figure~\ref{fig:zone_lose}. As there is no reference for neighborhood boundary like $b^k$ in $\mathbb{D}_{win}$, the breadth of the neighborhood $\delta_{lose}$ is a hyper-parameter set according to the dataset. As such, the right boundary $p_{lose}^r = b^k + \delta_{lose}$. The third loss function is devised as follows:
\begin{equation}
    \begin{aligned}
        Loss_3
        &= -\log\prod_{(\boldsymbol{x}^k,z^k)\in\mathbb{D}_{lose}} \int_{U^{+}(b^k,\delta_{lose})} pr_z(p|\boldsymbol{x}^k) \mathrm{d}p\\
        &= -\log\prod_{(\boldsymbol{x}^k,z^k)\in\mathbb{D}_{lose}} \sum_{i=i_{b^k}+1}^{i_{p_{lose}^r}} pr_i(i|\boldsymbol{x}^k)\\
        &\approx -\sum_{(\boldsymbol{x}^k,z^k)\in\mathbb{D}_{lose}} \log\sum_{i=i_{b^k}+1}^{i_{p_{lose}^r}} pr_i(i|\boldsymbol{x}^k)\\
    \end{aligned}
\end{equation}

Finally, these three losses are merged by weighted sum as follows:
\begin{equation}
    \begin{aligned}
        Loss_{nll} = \alpha \times Loss_1 + (1 - \alpha) \times (\beta \times Loss_2 + (1 - \beta) \times Loss_3)
    \end{aligned}
\end{equation}
where $\alpha$ and $\beta$ are hyper-parameters. 

Compared with setting a fixed zone breadth, NLL has different breadths for each sample in training, which describes the landscape more accurately. And compared to the losses defined in previous works, NLL introduces neighborhoods and divides optimization zones more accurately in comparison to roughly using $b^k$ and $z^k$. This loss avoids optimizing objectives on prices that are either too low or too high prices, reducing much of the error. When the buckets are narrow enough, the "approximately equal" in the above formulas can be ignored. And the number of buckets will not impact the efficiency of the model with the framework introduced in the next section.

\subsubsection{Model Structure}

\begin{figure}[ht]
  \centering
  \includegraphics[width=\linewidth]{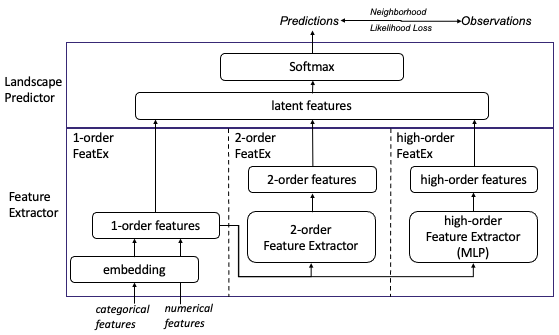}
  \caption{The structure of Arbitrary Distribution Modeling framework.}
  \Description{The structure of Arbitrary Distribution Modeling framework.}
  \label{fig:adm}
\end{figure}

Because in the RTB scenario, ad server execution and response have strict Service Level Agreement (SLA), we transfer more modeling responsibility and time complexity to offline training through NLL, and avoid designing complex structures for online model. The framework is consists of two tiers: Feature Extractor and Landscape Predictor, as Figure~\ref{fig:adm} shows,

Feature Extractor ($FeatEx$) maps the original features from the raw feature space into the latent factor space. Previous works have mostly focused on how to leverage the censored data in the loss function to better model the landscape but have overlooked the relationship between features, which plays an important role in characterizing the inventory\cite{mywork,ncf,nfm}. The original categorical features $\boldsymbol{x}^k_{cate}$ are embedded and concatenated with the normalized numerical features $\boldsymbol{x}^k_{num}$ directly to form the 1-order feature vector $\boldsymbol{x}^k_1$. The 2-order features $\boldsymbol{x}^k_2$ are extracted from $\boldsymbol{x}^k_1$ by the 2-order Feature Extractor $FeatEx_2(*)$, which could be any advanced feature interaction components in CTR prediction task, like FM\cite{fm}, CIN\cite{cin}, AutoInt\cite{autoint}, etc. For the high-order features, we stack a set of fully connected layers to extract the high-order feature combination $\boldsymbol{x}^k_h$ for its effective-efficiency balance \cite{mlp1, mlp2, mlp3, mlp4}. This process can be expressed as Equation~\ref{equ:FeatEx} shows,
\begin{equation}
\label{equ:FeatEx}
\begin{aligned}
    \boldsymbol{x}^k_1 &= FeatEx_1(\boldsymbol{x}^k) = [\boldsymbol{D}_{emb}\boldsymbol{x}^k_{cate}, norm(\boldsymbol{x}^k_{num})] \\
    \boldsymbol{x}^k_2 &= R_2(FeatEx_2(R_1(\boldsymbol{x}^k_1))) \\
    \boldsymbol{x}^k_{h_l} &= FeatEx_h(\boldsymbol{x}^k_1) = a_l(\boldsymbol{W}_l\boldsymbol{x}^k_{h_l}+\boldsymbol{b}_l) \\
    \boldsymbol{x}^k_{h_i} &=  a_i(\boldsymbol{W_i}\boldsymbol{x}^k_{h_i}+\boldsymbol{b_i}), i = 1, \dots, l, x_{h_0}^k = x_1^k 
\end{aligned}
\end{equation}
where $\boldsymbol{x}^k_{cate}$ is the one-hot encoding vector of the original categorical features and $\boldsymbol{D}_{emb}$ is the embedding dictionary. The $norm(*)$ function is the normalization operation applied on the original numerical features in the data pre-processing stage. The $R_1(*)$ function is used to reshape $\boldsymbol{x}^k_1$ into the correct shape that $FeatEx_2(*)$ component requires, and $R_2(*)$ flattens the output for further fusion with other feature vectors. The $a_i(*)$, $\boldsymbol{W}_i$ and $\boldsymbol{b}_i$ denote the activation function, the weights and bias parameters of the $i$-th fully connected layers of the Multi-Layer Perceptron (MLP), and $l$ is the number of the layers in $FeatEx_h(*)$.

The Landscape Predictor concatenates the 1-order, 2-order, and high-order feature vectors and feeds it into a Softmax layer to predict the probabilities of the winning price falling into each price bucket, as Equation~\ref{equ:LanPred} shows,
\begin{equation}
\label{equ:LanPred}
\begin{aligned}
    \boldsymbol{pr}^k = Softmax([\boldsymbol{x}^k_1,\boldsymbol{x}^k_2,\boldsymbol{x}^k_h])
\end{aligned}
\end{equation}
where $\boldsymbol{pr}^k$ is a vector with the length of the number of buckets, indicating the distribution of the winning price within the price range for ad request $\boldsymbol{x}^k$. 

More specifically, with the Equations above and all the parameters in ADM denoted as $\boldsymbol{\theta}$, the probability of one price value $p$ to be the winning price under the condition $\boldsymbol{x}^k$ is calculated as follow
\begin{equation}
\label{equ:Pred}
\begin{aligned}
    P(z^k=p|\boldsymbol{x}^k) &= pr_z(p|\boldsymbol{x}^k) \approx pr_i(i_p|\boldsymbol{x}^k) \\
     &= \boldsymbol{pr}^k[i_p] = \frac{e^{\boldsymbol{w}_{i_p}[\boldsymbol{x}^k_1,\boldsymbol{x}^k_2,\boldsymbol{x}^k_h]^T}}{\sum_{n=0}^{N-1}e^{\boldsymbol{w}_{n}[\boldsymbol{x}^k_1,\boldsymbol{x}^k_2,\boldsymbol{x}^k_h]^T}}
\end{aligned}
\end{equation}
where $i_p$ denotes the interval index of the price value $p$ and $\boldsymbol{pr}^k[i_p]$ is the $i_p$-th element in $\boldsymbol{pr}^k$.

\section{Experiments}
\label{page:experiments}

\subsection{Datasets}
We used two commonly-used public datasets, \href{https://contest.ipinyou.com}{iPinYou} \cite{ipinyou1,ipinyou2} and \href{http://apex.sjtu.edu.cn/datasets/7}{YOYI} \cite{yoyi1,yoyi2}, to evaluate whether ADM reaches state-of-the-art performance. Because these datasets do not contain the winning prices of the losing bids, we followed the common practice to drop the losing bids and fake bidding prices to simulate the winning and losing samples \cite{mixture1,dlf}, although many works have verified that the distributions of the winning prices in the winning bids and the losing bids are different \cite{mixture1}. We have also confirmed this as shown in Figure~\ref{fig:dist_z}:
\begin{figure}[ht]
\centering
    \subfigure[The distribution of the winning price of the bids one DSP attended.]{
    \centering
        \includegraphics[scale=0.25]{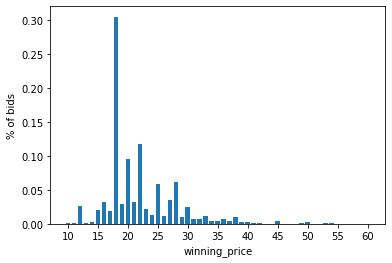}
        \label{fig:dist_z_all}
    }
    \subfigure[The distribution of the winning price of the bids one DSP won.]{
    \centering
        \includegraphics[scale=0.25]{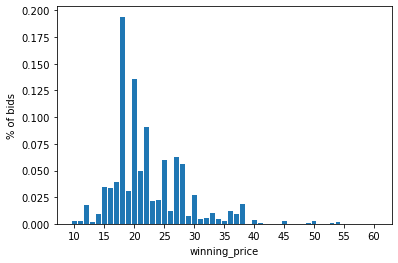}
        \label{fig:dist_z_win}
    }
\caption{The distribution of the winning price (The price values in the x-axis are scaled).}
\Description{The distribution of the winning price.}
\label{fig:dist_z}
\end{figure}

Thus, in our work, we also conducted experiments on the data extracted from our system. Our ad exchange server can observe the winning price of each auction, that is, the winning price of losing bids can also be acquired. Our methodology is to choose one representative DSP and collect all RTB auctions it participated in. The winning prices of its losing bids can be included in our dataset, but only to be used for evaluation, not for training. The evaluation on the true distribution of the winning prices makes our conclusion more convincing and reliable. 

The feature fields used in our dataset includes:
\begin{itemize}
    \item publisher features: \textit{URL, video\_group,  ad\_position}, etc.
    \item user features: \textit{country, city, device\_type, platform\_type}, etc.
    \item ad features: \textit{brand\_id, duration}, etc.
    \item context features: \textit{utc\_hour, day\_of\_week}, etc.
\end{itemize}
All the features are hashed and desensitized in compliance with corresponding rules and regulations to protect data privacy. The categorical features are encoded to indices for embedding. The embedding size follows the rule of thumb of taking the quadratic root of the cardinality of the feature. The missing values are treated as a single category. The numerical features are normalized by Min-Max Normalization and the missing values are filled in with the weighted average. For the price value, we divided the original price by the duration to get the unit price and took the logarithm of the unit price to scale the price range. We found that 99.8\% of the unit log price dropped into the range $[-3.0, 4.0]$, so we set the price bucket interval to 0.1 and obtained 70 price intervals.

All the auctions the DSP participated in within a day were used as the training dataset, which amounts to over 21 million bids, excluding about 10,000 bids that were randomly sampled as the validation dataset. The test dataset has about 10,000 randomly sampled bids from all the auctions the DSP attended the next day. The data was reformatted to be suitable for each baseline.

\subsection{Baselines and Settings}

The following models were selected as the baselines and fine-tuned with grid search strategy on each dataset to get their best performance for fair performance comparison. Due to the space constraint, their best settings on the three datasets are omitted.
\begin{itemize}
    \item \textbf{Dataset}: we present the original DSP bidding results in the dataset. Note that its price decision process included extra business rules and constraints.
    \item \textbf{Random model (RDM)}: This is a naive model that randomly bids the price within the price range.
    \item \textbf{Average model (AVG)}: This is a naive model that simply bids the average winning price of the winning bids in the training dataset.
    \item \textbf{Frequency model (FRQ)}: This is a naive model that bids based on the distribution of the observed winning prices. FRQ can be considered an advanced Random model as RDM samples bidding prices uniformly.
    \item \textbf{Censorship Linear Model (CLM)}: This model makes a point estimation based on the 1-order feature combination with both the winning and censored samples \cite{mixture1}. To compare it with other probabilistic models, we employed the Normal distribution on it as its prior distribution form. We set the prediction as the mean and fine-tuned the variance for each dataset to get its best performance.
    \item \textbf{Mixture model (MIX)}: This model is proposed by \cite{mixture1}, which is based on the Normal distribution assumption, with the variance as a hyper-parameter and directly predicting the mean value. It showed significant improvements compared to previous works. As normal distribution is the most commonly used assumption, we take this model as a representative of pre-assumed probabilistic estimators.
    \item \textbf{Deep Landscape Forecasting (DLF)}: This model is proposed by \cite{dlf}. It leverages conditional probability chain rule and LSTM to model the landscape and get rid of prior assumptions. The price with the highest probability is taken as the prediction. This model achieved remarkable results on distribution modeling. As such, we will take this model to be a state-of-the-art baseline of probabilistic models. The data of each dataset was reformatted to time-series format which is suitable for this model.
\end{itemize}

The settings of ADM are as follow. For $FeatEx_2$, we tried several advanced feature interaction components in CTR prediction models: InnerProduct in PNN (InnerProd) \cite{innerprod}, Factorization Machine (FM) \cite{fm}, Attention Factorization Machine (AFM) \cite{afm}, the Cross Network part in Deep \& Cross Network - Mix (DCN-Mix) \cite{dcn-mix}, Compressed Interaction Network in xDeepFM (CIN) \cite{cin}, Interacting Layer in AutoInt (AutoInt) \cite{autoint}, Bi-Interaction Layer in Neural Factorization Machine (NFM) \cite{nfm}, and Collaborative Filtering (CF) \cite{ncf}. The 1-order features $\boldsymbol{x}=\{\boldsymbol{x}_p,\boldsymbol{x}_u,\boldsymbol{x}_a,\boldsymbol{x}_c\}$ is directly fed into these components to get 2-order features. Because this is not our focus in this paper, we will not repeat the details and the distinction of these algorithms here. For $FeatEx_h$, the activation functions of each layer were set to ReLU and the number of layers of MLP to 3, with the size of each layer being half of the former. 

For the hyper-parameters in NLL, $r_{win}^l$ and $r_{win}^l$ were set to $1$. $\delta_{lose}$ was set to 40, $\alpha$ was set to 0.2 and $\beta$ was set to 0.8 after tuning.

All the experiments were performed on 8 $\times$ AWS ml.c5.18xlarge EC2 instances with sufficient resources to fairly compare their time efficiency. 

\subsection{Metrics}

We evaluated the performance of the baselines and ADM from these perspectives:

\subsubsection{Regression Metrics} Predicting the winning price is ultimately a regression task, so the accuracy of the final result is important. The price with the highest probability in the prediction of a probabilistic model, denoted as $\hat{z}^k$, is selected as the winning price prediction. We used Mean Average Error (\textbf{MAE}) between the prediction $\hat{z}^k$ and the ground-truth $z^k$, to assess the model's performance on regression accuracy, as Equation~\ref{equ:mae} shows.
\begin{equation}
    \begin{aligned}
        MAE = \frac{1}{\left|\mathbb{D}_{test}\right|} \sum_{(\boldsymbol{x}^k,z^k)\in\mathbb{D}_{test}} |\hat{z}^k - z^k|
        \label{equ:mae}
    \end{aligned}
\end{equation}
  
\subsubsection{Probabilistic Modeling Metrics} As a probabilistic model, the capability of modeling a distribution should be considered. We use two commonly-used metrics \cite{dlf}, Average Negative Log Probability (\textbf{ANLP}) and Concordance Index (\textbf{C-Index}). 

ANLP is to assess the likelihood of the winning prices as Equation~\ref{equ:anlp}. A lower ANLP is better because it indicates consistency with the actual situation.
\begin{equation}
\begin{aligned}
ANLP = - \frac{1}{\left|\mathbb{D}_{test}\right|} \sum_{(\boldsymbol{x}^k,z^k)\in\mathbb{D}_{test}} \log pr_z(z^k|\boldsymbol{x}^k)
\label{equ:anlp}
\end{aligned}
\end{equation}

C-Index is used to measure how well the model is ordering the samples based on the winning prices. For instance, for any sample $(\boldsymbol{x}^{k_1}, z^{k_1})$, samples $(\boldsymbol{x}^{k_2}, z^{k_2})$ with $z^{k_2} > z^{k_1}$ should always be placed in front of $(\boldsymbol{x}^{k_1}, z^{k_1})$. This means that for a specific price value, the C-Index assesses how well the model manages to place the winning bids before the losing bids. This is the same as the area under the ROC curve (AUC) metric in classification tasks. We regarded the winning or losing of one bid as the label and the winning rate of the winning price $wr(z^k)$ as the class confidence, and then calculated the AUC metric. Therefore, the C-Index illustrates its probabilistic modeling performance.

\subsubsection{Business Metrics} Besides the algorithm metrics, business performance is vital and indicates whether it's a \textit{good} algorithm to release to production. We use the number of wins and the value of wins to describe the changes of ad campaigns. As a DSP in an RTB auction, we always aim to win as much as possible \cite{beeswax}. As such, the \textbf{number of wins} is an important business metric. In addition to quantity, quality is also an important indicator that cannot be ignored. High Click Through Rate (CTR) or high Order Conversion Rate (OCR) ad impressions are considered as high quality and valuable inventory, so their prices are relatively higher. Therefore, we also measured the \textbf{Value} of one set of bids by its average winning price since the winning price indicates the market's recognition of the value of the bid. As such, in addition to winning as much as possible, we also aimed to obtain a set of high-value winning bids.

Finally, besides the effectiveness, we also focused on the efficiency of our algorithm. The training time in offline and inference latency in online serving are tested on the production data by replaying the data of the day.

\section{Evaluation}
\label{page:evaluation}

In this section, we evaluated their performance on two public datasets and our production dataset. For our dataset, all the prices have been converted to unit-duration CPM in dollars. And it is noteworthy that although the difference in the number of wins between baselines is not so conspicuous, considering that this test dataset was only a 0.46\textperthousand sampling of the daily traffic, it may win a lot more in the long run.

\subsection{Performance on Public Datasets}

We compared ADM with state-of-the-art algorithms on iPinYou and YOYI datasets with the same metrics used in their papers. As the results of CLM and MIX were not comparable to DLF and ADM, they are not presented in Table~\ref{tab:anlp} and Table~\ref{tab:c-index}. For ANLP, ADM outperforms DLF both on iPinYou and YOYI. For C-Index, ADM reached the same effects as DLF.
\begin{table}
  \begin{minipage}[t]{0.2\textwidth}
  \centering
  \caption{ANLP on iPinYou and YOYI}
  \label{tab:anlp}
  \begin{tabular}{ccc}
    \toprule
    Advertiser & DLF & ADM \\
    \midrule
    1458 & 4.088 & \textbf{3.842} \\
    2259 & 5.244 & \textbf{4.687} \\
    2261 & 4.632 & \textbf{3.723} \\
    2821 & 5.428 & \textbf{4.653} \\
    2997 & 4.504 & \textbf{3.786} \\
    3358 & 5.281 & \textbf{4.561} \\
    2286 & 4.940 & \textbf{3.863} \\
    3427 & 4.836 & \textbf{3.918} \\
    3476 & 4.012 & \textbf{0.922} \\
    Overall & 4.774 & \textbf{4.100} \\
    \midrule
    YOYI & 4.453 & \textbf{3.520} \\
    \bottomrule
  \end{tabular}
  \end{minipage}
  \hspace{0.5cm} 
  \begin{minipage}[t]{0.2\textwidth}
  \centering
  \caption{C-Index on iPinYou and YOYI}
  \label{tab:c-index}
  \begin{tabular}{ccc}
    \toprule
    Advertiser & DLF & ADM \\
    \midrule
    1458 & 0.904 & \textbf{0.911} \\
    2259 & 0.876 & \textbf{0.894} \\
    2261 & \textbf{0.929} & 0.921 \\
    2821 & 0.881 & \textbf{0.900} \\
    2997 & 0.919 & \textbf{0.930} \\
    3358 & \textbf{0.944} & 0.867 \\
    2286 & \textbf{0.923} & 0.908 \\
    3427 & 0.901 & \textbf{0.905} \\
    3476 & \textbf{0.922} & 0.920 \\
    Overall & \textbf{0.911} & 0.906 \\
    \midrule
    YOYI & 0.924 & \textbf{0.944}\\
    \bottomrule
  \end{tabular}
  \end{minipage}
\end{table}

\subsection{Effectiveness of NLL}

\begin{table}
  \caption{Results of Baselines w/ or w/o NLL}
  \label{tab:nll}
  \begin{tabular}{cccccc}
    \toprule
    Baselines & MAE & ANLP & C-Index & $\sharp$ of wins & Value\\
    \midrule
    Dataset & 0.26 & - & - & 3795 & 1.03 \\
    \midrule
    RDM & 0.53 & - & - & 5638 & 0.80 \\
    AVG & 0.43 & - & - & 5302 & 0.67 \\
    FRQ & 0.52 & - & - & 4991 & 0.77 \\
    CLM & 0.57 & \textbf{3.65} & 0.81 & 7458 & 0.86 \\
    MIX & 0.55 & 3.70 & 0.79 & 7565 & 0.88 \\
    DLF & 0.51 & 3.85 & \textbf{0.92} & 8006 & \textbf{0.96} \\
    \midrule
    DLF (NLL) & 0.42 & 3.95 & \textbf{0.92} & \textbf{8155} & \textbf{0.96} \\
    ADM (NLL) & \textbf{0.36} & 3.66 & 0.90 & 8035 & \textbf{0.96} \\
    \bottomrule
\end{tabular}
\end{table}

We replaced the original losses of DLF with NLL and denoted it as DLF (NLL). We also utilized the simplest ADM structure - one hidden layer to extract the latent factor with no $FeatEx_2$ and $FeatEx_h$, denoted as ADM (NLL).

Table~\ref{tab:nll} is the results of the experiments of NLL. The best value is shown in bold. For the regression performance, although ADM in this experiment didn't leverage any sophisticated structure, it still outperformed other non-NLL models at least $16\%$ on MAE. For the probabilistic modeling performance, ADM reached almost the best ANLP and the two models with NLL reached state-of-the-art performance \cite{dlf} on C-Index along with DLF. Generally, an estimator asks for a trade-off between regression accuracy and probabilistic modeling capability. The results demonstrate that NLL can improve the accuracy while keeping the powerful capability of probabilistic modeling. This is owed to the accurate promotion zone of NLL. Last but not least, for the business metrics, the models with NLL also outperformed the others by winning the most bids, which were also the most valuable sets of winning bids. Therefore, we can see the advantages of NLL on predicting accuracy, probabilistic modeling, as well as business profits.

\subsection{Effectiveness of High-Level Interaction}

\begin{table}
  \caption{Results of Baselines with Different 2-Order Interaction Components}
  \label{tab:2-order}
  \begin{tabular}{cccccc}
    \toprule
    $FeatEx_2$ & MAE & ANLP & C-Index & $\sharp$ of wins & Value\\
    \midrule
    W/O & 0.36 & 3.66 & 0.90 & 8035 & \textbf{0.96} \\
    InnerProd & 0.35 & 4.63 & 0.87 & 7860 & 0.94 \\
    FM & 0.37 & 4.70 & 0.88 & 7969 & 0.95 \\
    AFM & 0.33& 4.64 & 0.87 & 7806 & 0.93 \\
    DCN-Mix & 0.33 & 3.66 & \textbf{0.91} & 8021 & \textbf{0.96} \\
    CIN & 0.33 & 3.71 & \textbf{0.91} & 8041 & \textbf{0.96} \\
    AutoInt & \textbf{0.32} & 3.62 & \textbf{0.91} & 8092 & \textbf{0.96} \\
    NFM & 0.34 & 3.73 & 0.90 & 8101 & \textbf{0.96} \\
    CF & 0.34 & \textbf{3.60} & 0.90 & \textbf{8131} & \textbf{0.96} \\
    \bottomrule
\end{tabular}
\end{table}

In this part, we leveraged various 2-order feature interaction components as $FeatEx_2$ and an MLP as $FeatEx_h$, comparing them with one without any high-level Feature Extractors (denoted as W/O) to demonstrate the best performance ADM can reach.

Table~\ref{tab:2-order} is the performance on different $FeatEx_2$. Compared to W/O, we found that most $FeatEx_2$ are helpful except for InnerProd, FM, and AFM. Almost all the components improved MAE and the Interacting Layer in AutoInt achieved the best score. Though these $FeatEx_2$ didn't make any significant improvements on the probability modeling capability, some of them did increase the number of winning bids, while keeping the value of the winning bids the same. The reason why InnerProd, FM, and AFM didn't perform well may be that the outputs of these components are single-element tensors, which weakens their capability of representation in the framework.

In summary, ADM is able to be introduced by various modules for different data distribution or task objectives. The naive ADM framework provides great potential in improvements and strong flexibility in customization.

\subsection{Hyper-parameters}

\subsubsection{Neighborhood Boundary}

\begin{figure}[ht]
  \centering
  \includegraphics[width=\linewidth]{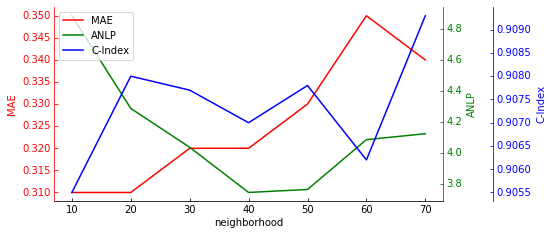}
  \caption{The performance of different neighborhood settings.}
  \Description{The performance of different neighborhood settings.}
  \label{fig:hp_nei}
\end{figure}

Since the breadth of optimization zone in $mathbb{D}_{win}$ is the ratio of $\delta_{win}$, and the breadth of optimization zone in $mathbb{D}_{lose}$ has no reference, the value range of $\delta_{lose}$ is wider, so it has a greater impact on the model performance. We studied the impact of $p_{lose}^r$ in $Loss_3$ on the performance. For each sample in $\mathbb{D}_{lose}$, $p_{lose}^r$ was set to the minimum value of a default setting and the length of the price range, which is 70 in our dataset. As Figure~\ref{fig:hp_nei} shows, MAE kept going up as the neighborhood increased because it gave a less strict promotion zone for winning price candidates. However, the modeling metrics showed better performance when they were in the middle of the range, which indicates that the performance gets better when the setting of $\delta_{lose}$ is closer to the true difference between $b$ and $z$ in $\mathbb{D}_{lose}$. In our experiments, we chose 40 as the right boundary of the neighborhood to balance the accuracy and modeling capability.

\subsubsection{Coefficients in NLL}

\begin{figure}[ht]
  \centering
  \includegraphics[width=\linewidth]{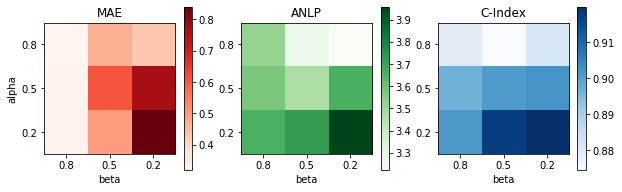}
  \caption{The performance of different $\alpha$ and $\beta$ settings.}
  \Description{The performance of different alpha and beta settings.}
  \label{fig:ab}
\end{figure}

We studied the performance of different coefficients to see the impact of each component in NLL. We conducted a series of experiments on $\alpha$ and $\beta$ from $[0.2, 0.5, 0.8]$. From the heat maps in Figure~\ref{fig:ab}, we can see that a large $\alpha$ is good for ANLP because $Loss_1$ is consistent with ANLP. Meanwhile, a large $\beta$ is good for predicting accuracy because it reinforces the weights of $Loss_2$, where the ground truth $z$ is used. As such, the model's coefficients can be adjusted as needed.

\subsection{Performance in System}

We replayed historical data of the production for days in our system, where there are over 7 million requests the chosen DSP responses daily. On average, ADM can win $63.55\%$ of the auctions the DSP participated in, and win $21.45\%$ more bids than the original strategy of the DSP. If we define the proportion of the number of bids that ADM and DSP both win against those that the DSP win to be \textit{Recall}, the score is $75\%$. Though the original strategy of the DSP won less, the value of its wins is about 1.06, compared to 0.93 for ADM, which indicates that the part ADM wins over is less valuable.

\begin{figure}
  \centering
  \includegraphics[width=\linewidth]{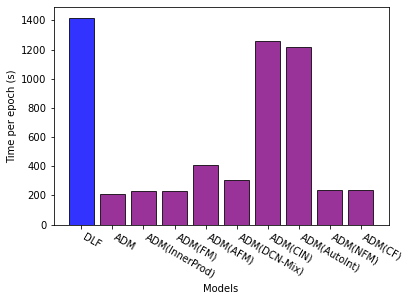}
  \caption{The time cost of different non-assumption probabilistic models in offline training.}
  \Description{The time cost of different non-assumption probabilistic models in offline training.}
  \label{fig:time_offline}
\end{figure}

The offline training time cost of DLF and ADM variants is shown in Figure~\ref{fig:time_offline}. Due to the temporal dependency of RNN, DLF spent over 1400s per epoch. This shortcoming becomes more severe if long time steps with narrow buckets are used to guarantee the prediction precision. ADM is essentially an MLP model (if no RNN components are involved), and it saved at most 85$\%$ time cost while keeping state-of-the-art performance. CIN and AutoInt had a large time cost due to their complex structures such as Convolutional Neural Networks (CNN), but this could be alleviated by using GPU instances.

\begin{figure}
  \centering
  \includegraphics[width=\linewidth]{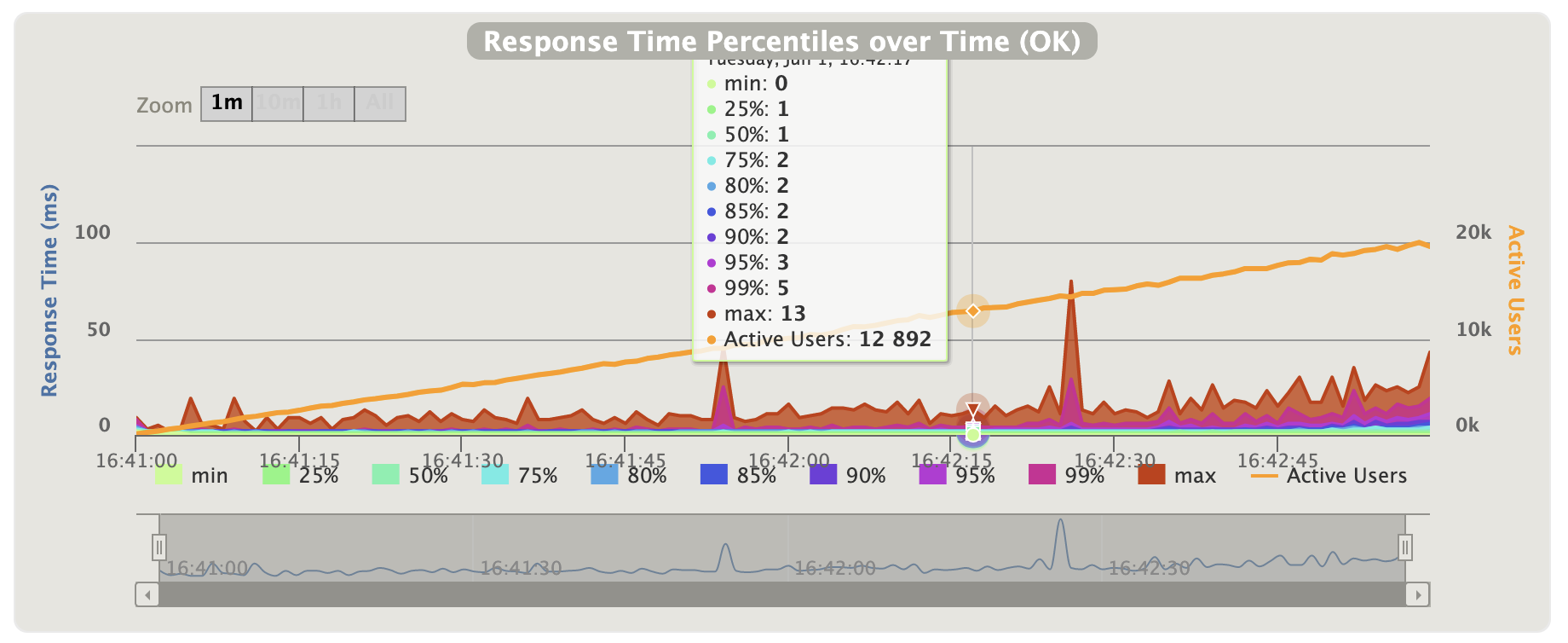}
  \caption{The online serving time cost of ADM in our system. (The peaks in the graph were caused by network jitter.)}
  \Description{The online serving time cost of ADM in our system.}
  \label{fig:time_online}
\end{figure}

The online serving time performance of ADM is shown in Figure~\ref{fig:time_online}. The graph presents the serving performance of 20k requests in 2 consecutive minutes under an ELB of 4 pods. Benefit from the concise structure of ADM, over $99\%$ of the response time is under 5ms, and according to the monitor the inference time cost is only around 3ms, which meets the SLA of the online system perfectly.

\section{Conclusion}
\label{page:conclusion}

In this paper, we proposed Neighborhood Likelihood Loss to solve the bidding landscape forecasting problem more accurately. We introduced this novel loss function into the Arbitrary Distribution Modeling framework, which makes no prior assumption of distribution form and executes efficiently in online service. ADM with NLL showed its advantages on both algorithm and business metrics on public and production datasets. It reached state-of-the-art performance with significantly less time cost and has been serving effectively and efficiently in our system.

For future work, the probabilities of the prices in the optimization zones could be promoted in weight. Also, as we merely studied the Feature Extractors and advanced model structures of ADM, there is great potential for improvements with a focus on the representation learning or bid shaping of the RTB auction. Furthermore, we didn't include any expected Key Performance Index (KPI) in this model, such as CTR or OCR, despite the fact that they are critical for online advertising. We expect that the performance of ADM could be further improved as they are integrated.

\bibliographystyle{ACM-Reference-Format}
\bibliography{www22}

\end{document}